\newcommand{\parskiny}{\vspace{0.1mm}}
\newcommand{\seckiny}{\vspace{0.1mm}}
\newcommand{\subseckiny}{\vspace{0.1mm}}
\newcommand{\figskiny}{\vspace{0.1mm}}

\newcommand{\done}[1]{}

\documentclass{article}

\usepackage{times}
\usepackage{graphicx} 
\usepackage{subfig}
\usepackage{adjustbox}
\usepackage{amsfonts}

\usepackage{natbib}

\usepackage{algorithm}
\usepackage{algorithmic}

\usepackage{hyperref}

\usepackage[accepted]{icml2016}

\icmltitlerunning{FeUdal Networks for Hierarchical Reinforcement Learning}

\begin{document} 

\twocolumn[
\icmltitle{FeUdal Networks for Hierarchical Reinforcement Learning}

\icmlauthor{Alexander Sasha Vezhnevets}{vezhnick@google.com}
\icmlauthor{Simon Osindero}{osindero@google.com}
\icmlauthor{Tom  Schaul}{schaul@google.com}
\icmlauthor{Nicolas  Heess}{heess@google.com}
\icmlauthor{Max  Jaderberg}{jaderberg@google.com}
\icmlauthor{David  Silver}{davidsilver@google.com}
\icmlauthor{Koray   Kavukcuoglu}{korayk@google.com}
\icmladdress{DeepMind}

\icmlkeywords{deep learning, reinforcement learning, hierarchical learning}

\vskip 0.3in
]

\begin{abstract}

We introduce FeUdal Networks (FuNs): a novel architecture for hierarchical reinforcement learning.
Our approach is inspired by the feudal reinforcement learning proposal of Dayan and Hinton, and gains power and efficacy by decoupling end-to-end learning across multiple levels -- allowing it to utilise different resolutions of time.
Our framework employs a Manager module and a Worker module. 
The Manager operates at a lower temporal resolution and sets abstract goals which are conveyed to and enacted by the Worker. The Worker generates primitive actions at every tick of the environment.
The decoupled structure of FuN conveys several benefits -- in addition to facilitating very long timescale credit assignment it also encourages the emergence of sub-policies associated with different goals set by the Manager. 
These properties allow FuN to dramatically outperform a strong baseline agent on tasks that involve long-term credit assignment or memorisation.
We demonstrate the performance of our proposed system on a range of tasks from the ATARI suite and also from a 3D DeepMind Lab environment.

\end{abstract}
\seckiny
\seckiny

\seckiny
\section{Introduction}
\seckiny

Deep reinforcement learning has recently enjoyed successes in many domains~\cite{mnih-dqn-2015,schulman2015trust,levine2015endtoend,mnih2016asynchronous,lillicrap2015continuous}. 
Nevertheless, long-term credit assignment remains a major challenge for these methods, especially in environments with sparse reward signals, such as the infamous Montezuma's Revenge ATARI game. 
It is symptomatic that the standard approach on the ATARI benchmark suite~\cite{bellemare-ale} is to use an action-repeat heuristic, where each action translates into several (usually $4$) consecutive actions in the environment. 
Yet another dimension of complexity is seen in non-Markovian environments that require memory -- these are particularly challenging, since the agent has to learn which parts of experience to store for later, using only a sparse reward signal.

The framework we propose takes inspiration from feudal reinforcement learning (FRL) introduced by \citet{dayan1993feudal}, where levels of hierarchy within an agent communicate via explicit goals. 
Some key insights from FRL are that goals can be generated in a top-down fashion, and that goal setting can be decoupled from  goal achievement; a level in the hierarchy communicates to the level below it what must be achieved, but does not specify \emph{how} to do so.  Making higher levels reason at a lower temporal resolution naturally structures the agents behaviour into temporally extended sub-policies.  

The architecture explored in this work is a fully-differentiable neural network with two levels of hierarchy (though there are obvious generalisations to deeper hierarchies).
The top level, the “Manager”, sets goals at a lower temporal resolution in a latent state-space that is itself \textit{learnt} by the Manager.
The lower level, the “Worker”, operates at a higher temporal resolution and produces primitive actions, conditioned on the goals it receives from the Manager. The Worker is motivated to follow the goals by an intrinsic reward.
However, significantly, no gradients are propagated between Worker and Manager; the Manager receives its learning signal from the environment alone. In other words, the Manager learns to select latent goals that maximise extrinsic reward.

The key contributions of our proposal are: (1) A consistent, end-to-end differentiable model that embodies and generalizes the principles of FRL. (2) A novel, approximate \textit{transition policy gradient} update for training the Manager, which exploits the semantic meaning of the goals it produces. (3) The use of goals that are directional rather than absolute in nature. (4) A novel RNN design for the Manager -- a dilated LSTM -- which extends the longevity of the recurrent state memories and allows gradients to flow through large hops in time, enabling effective back-propagation through hundreds of steps.

Our ablative analysis (Section \ref{subsec:ablative}) confirms that transitional policy gradient and directional goals are crucial for best performance. Our experiments on a selection of ATARI games (including the infamous Montezuma's revenge) and on several memory tasks in the 3D DeepMind Lab environment~\cite{beattie2016labyrinth} show that FuN significantly improves long-term credit assignment and memorisation. 

\seckiny

\section{Related Work}
\label{sec:related_work}
\seckiny

Building hierarchical agents is a long standing topic in reinforcement learning~\cite{sutton1999between,precup2000temporal, dayan1993feudal, dietterich2000hierarchical, boutilier1997prioritized, dayan1993improving, kaelbling2014hierarchical, parr1998reinforcement, precup1997planning, precup1998theoretical, schmidhuber1991neural, sutton1995td, wiering1997hq,vezhnevets2016straw,bacon2015option}. 
The options framework~\cite{sutton1999between,precup2000temporal} is a popular formulation for considering the problem with a two level hierarchy. The bottom level -- an option -- is a sub-policy with a termination condition, which takes in environment observations and outputs actions until the termination condition is met. 
An agent picks an option using its policy-over-options (the top level) and subsequently follows it until termination, at which point the policy-over-options is queried again and the process continues. 
Options are typically learned using sub-goals and `pseudo-rewards' that are provided explicitly~\cite{sutton1999between, dietterich2000hierarchical, dayan1993feudal}. For a simple, tabular case~\cite{wiering1997hq,schaul2015uvfa}, each state can be used as a sub-goal.
Given the options, a policy-over-options can be learned using standard techniques by treating options as actions. 
Recently~\cite{tessler2016minecraft, tejas2016hdrl} have demonstrated that combining deep learning with pre-defined sub-goals delivers promising results in challenging environments like Minecraft and Atari, however sub-goal discovery was not addressed.

A recent work of~\cite{bacon2015option} shows the possibility of learning options jointly with a policy-over-options in an end-to-end fashion by extending the policy gradient theorem to options. 
When options are learnt end-to-end, they tend to degenerate to one of two trivial solutions: (i) only one active option that solves the whole task; (ii) a policy-over-options that changes options at every step, micro-managing the behaviour. Consequently, regularisers~\cite{bacon2015option,vezhnevets2016straw} are usually introduced to steer the solution towards multiple options of extended length. This is believed to provide an inductive bias towards re-usable temporal abstractions and to help generalisation. 

A key difference between our approach and the options framework is that in our proposal the top level produces a meaningful and explicit goal for the bottom level to achieve. Sub-goals emerge as directions in the latent state-space and are naturally diverse. We also achieve significantly better scores on ATARI than Option-Critic (section~\ref{sec:Experiments}).

There has also been a significant progress in non-hierarchical deep RL methods by using auxiliary losses and rewards. \cite{bellemare2016countexplore} have significantly advanced the state-of-the-art on Montezuma's Revenge by using pseudo-count based auxiliary rewards for exploration, which stimulate agents to explore new parts of the state space. The recently proposed UNREAL agent~\cite{jaderberg2016unreal} also demonstrates a strong improvement by using unsupervised auxiliary tasks to help refine its internal representations. We note that these benefits are orthogonal to those provided by FuN, and that both approaches could be combined with FuN for even greater effect.

\seckiny

\begin{figure}
    \centering
    \includegraphics[scale=0.4]{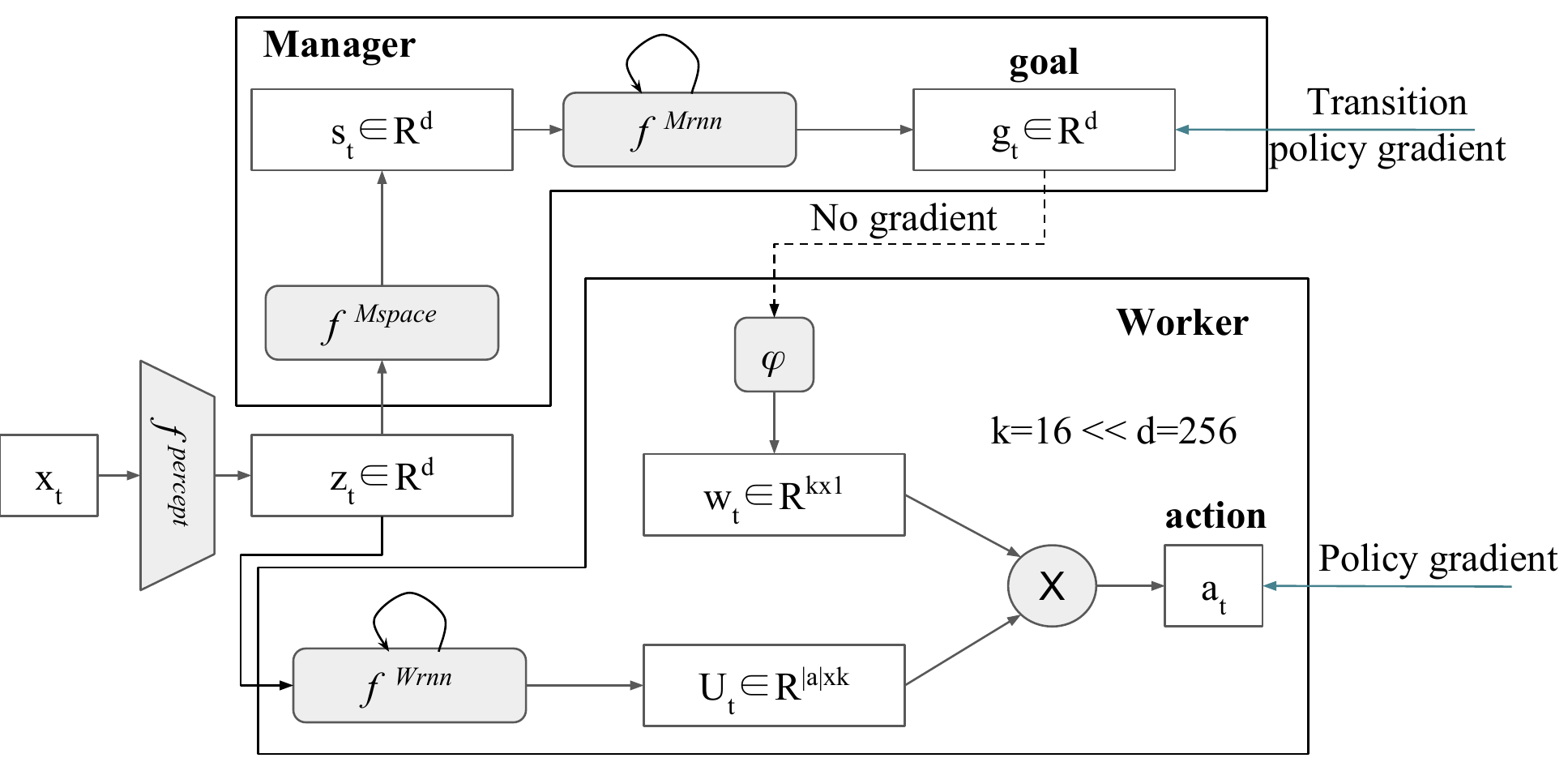}
    \caption{The schematic illustration of FuN (section~\ref{sec:model}) \figskiny }
    \label{fig:architecture}
\end{figure}

\section{The model}
\label{sec:model}

\paragraph{What is FuN?} 
FuN is a modular neural-network consisting of two modules -- the Worker and the Manager. The Manager internally computes a latent state representation $s_t$ and outputs a goal vector $g_t$. The Worker produces actions conditioned on external observation, it’s own state, and the Manager’s goal. The Manager and the Worker share a perceptual module which takes an observation from the environment $x_t$ and computes a shared intermediate representation $z_t$. 
The Manager's goals $g_t$ are trained using an approximate \textit{transition policy gradient}. This is a particularly efficient form of policy gradient training that exploits the knowledge that the Worker's behaviour will ultimately align with the goal directions it has been set. The Worker is then trained via intrinsic reward to produce actions that cause these goal directions to be achieved. Figure~\ref{fig:architecture}a illustrates the overall design and the following equations describe the forward dynamics of our network:

\begin{equation}
z_t = f^\textnormal{percept}(x_t)
\end{equation}
\begin{equation}
s_t = f^\textit{Mspace}(z_t)
\end{equation}
\begin{equation}
h^M_t, \hat{g}_t = f^\textit{Mrnn}(s_t, h^M_{t-1}); g_t=\hat{g}_t/||\hat{g}_t||; 
\end{equation}
\begin{equation}
w_t = \phi(\sum_{i=t-c}^t {g_i}) 
\label{eqn:FuN:pool_embed}
\end{equation}
\begin{equation}
h^W, U_t = f^\textit{Wrnn}(z_t, h^W_{t-1}) 
\label{eqn:FuN:psi}
\end{equation}
\begin{equation}
\pi_t = \textit{SoftMax}(U_t w_t)
\label{eqn:FuN:pi}
\end{equation}

where both the Manager and the Worker are recurrent. Here $h^M$ and $h^W$ correspond to the internal states of the Manager and the Worker respectively. A linear transform $\phi$ maps a goal $g_t$ into an embedding vector $w_t \in R^k$, which is then combined via product with matrix $U_t$ (Workers output) to produce policy $\pi$ -- vector of probabilities over primitive actions. 
The next section provides the details on goal embedding and the following sections~\ref{sec:Learning},\ref{subsec:tpg} describes how FuN is trained.

\subseckiny
\subsection{Goal embedding}
\subseckiny
\label{subsec:goal_embedding}
The goal $g$ modulates the policy via a multiplicative interaction in a low dimensional goal-embedding space $R^k, k<<d$. The Worker first produces an embedding vector for every action, represented by rows of matrix $U \in R^{|a| \times k}$ (eq.~\ref{eqn:FuN:psi}). 
To incorporate goals from the Manager, the last $c$ goals are first pooled by summation and then embedded into a vector $w \in R^k$ using a linear projection $\phi$ (eq.~\ref{eqn:FuN:pool_embed}). The projection $\phi$ is linear, with no biases, and is learnt with gradients coming from the Worker's actions. The embedding matrix $U$ is then combined with the goal embedding $w$ via a matrix-vector product (eq.~\ref{eqn:FuN:pi}).
Since $\phi$ has no biases it can never produce a constant non-zero vector  -- which is the only way the setup could ignore the Manager's input. This makes sure that the goal output by the Manager always influences the final policy. Notice how, due to pooling of goals over several time-steps, the conditioning from the Manager varies smoothly. 

\seckiny

\subsection{Learning}
\subseckiny
\label{sec:Learning}

We consider a standard reinforcement learning setup. At each step $t$, the agent receives an observation $x_t$ from the environment and selects an action $a_t$ from a finite set of possible actions. The environment responds with a new observation $x_{t+1}$ and a scalar reward $r_t$. The process continues until the terminal state is reached, after which it restarts. The goal of the agent is to maximise the discounted return 
$R_t= \sum_{k=0}^{\infty} \gamma^k r_{t+k+1}$ with $\gamma \in [0,1]$. The agent's behaviour is defined by its action-selection policy $\pi$. FuN produces a distribution over possible actions (a stochastic policy) as defined in eq.~\ref{eqn:FuN:pi}.

The conventional wisdom would be to train the whole architecture monolithically through gradient descent on either the policy directly or via TD-learning. Notice, that since FuN is fully differentiable we could train it end-to-end using a policy gradient algorithm operating on the actions taken by the Worker. The outputs $g$ of the Manager would be trained by gradients coming from the Worker. This, however would deprive Manager's goals $g$ of any semantic meaning, making them just internal latent variables of the model. 
We propose instead to independently train Manager to predict advantageous directions (transitions) in state space and to intrinsically reward the Worker to follow these directions.
If the Worker can fulfil the goal of moving in these directions (as it is rewarded for doing), then we ought to end up taking advantageous trajectories through state-space.
We formalise this in the following update rule for the Manager:

\begin{equation}
\nabla g_t = A^M_t \nabla_{\theta} d_{\cos}(s_{t+c} - s_t, g_t(\theta)),
\label{eqn:manager_update}
\end{equation}

where $A^M_t = R_t - V^M_t(x_t,\theta)$ is the Manager's advantage function, computed using a value function estimate $V^M_t(x_t,\theta)$ from the internal critic; $d_{\cos}(\alpha,\beta)=\alpha^T\beta/(|\alpha||\beta|)$ is the cosine similarity between two vectors. Note: the dependence of $s$ on $\theta$ is ignored when computing $\nabla_{\theta} d_{\cos}$ -- 
this avoids trivial solutions. Notice that now $g_t$ acquires a semantic meaning as an advantageous direction in the latent state space at a horizon $c$, which defines the temporal resolution of the Manager.

The intrinsic reward that encourages the Worker to follow the goals is defined as: 
\begin{equation}
r^I_t = 1/c \sum_{i=1}^c d_{\cos}(s_{t} - s_{t-i}, g_{t-i}) 
\label{eqn:intrinsic_reward}
\end{equation}

We use directions because it is more feasible for the Worker to be able to reliably cause directional shifts in the latent state than it is to assume that the Worker can take us to (potentially) arbitrary new absolute locations. It also gives a degree of invariance to the goals and allows for structural generalisation -- the same directional sub-goal $g$ can invoke a sub-policy that is valid and useful in a large part of the latent state space; e.g. evade an enemy, swim up for air, etc. We compare absolute against directional goals empirically in section~\ref{subsec:ablative}. 

The original feudal reinforcement learning formulation of~\citet{dayan1993feudal} advocated completely concealing the reward from the environment from lower levels of hierarchy. In practice we take a softer approach by adding an intrinsic reward for following the goals, but retaining the environment reward as well. The Worker is then trained to maximise a weighted sum $R_t + \alpha R^I_t$, where $\alpha$ is a hyper-parameter that regulates the influence of the intrinsic reward. The Worker’s policy $\pi$ can be trained to maximise intrinsic reward by using any off-the shelf deep reinforcement learning algorithm. Here we use an advantage actor critic~\cite{mnih2016asynchronous}:
\begin{equation}
\nabla \pi_t  = A^D_t \nabla_{\theta}\log\pi(a_t|x_t;\theta)
\label{eqn:doer_update}
\end{equation}

The Advantage function $A^D_t = (R_t + \alpha R^I_t - V^D_t(x_t;\theta))$
is calculated using an internal critic, which estimates the value functions for both rewards.

Note that the Worker and Manager can potentially have different discount factors $\gamma$ for computing the return. 
This allows, for instance, the Worker to be more greedy and focus on immediate rewards while the Manager can consider a long-term perspective.

\subseckiny
\subsection{Transition Policy Gradients}
\subseckiny
\label{subsec:tpg}

We now motivate our proposed update rule for the Manager as a novel form of policy gradient with respect to a \emph{model} of the Worker's behaviour.
Consider a high-level policy $o_t = \mu(s_t, \theta)$ that selects among sub-policies (possibly from a continuous set), where we assume for now that these sub-policies are fixed duration behaviours (lasting for $c$ steps). 
Corresponding to each sub-policy is a transition distribution, $p(s_{t+c}|s_t,o_t)$, that describes the distribution of states that we end up at the end of the sub-policy, given the start state and the sub-policy enacted. The high-level policy can be composed with the transition distribution to give a `transition policy' $\pi^{TP}(s_{t+c}|s_t) = p(s_{t+c}|s_t,\mu(s_t, \theta))$ describing the distribution over end states given start states. It is valid to refer to this as a policy because the original MDP is isomorphic to a new MDP with policy $\pi^{TP}$ and transition function $s_{t+c} = \pi^{TP}(s_t)$ (i.e. the state always transitions to the end state picked by the transition policy). As a result, we can apply the policy gradient theorem to the transition policy $\pi^{TP}$, so as to find the performance gradient with respect to the policy parameters, \vspace{-5mm}

\begin{equation}
\nabla_{\theta} \pi^{TP}_t = \mathbb{E} \left[ (R_t - V(s_t)) \nabla_{\theta} \log p(s_{t+c}|s_t, \mu(s_t,\theta))\right]
\label{eqn:tpg}
\end{equation}

In general, the Worker may follow a complex trajectory. A naive application of policy gradients requires the agent to learn from samples of these trajectories. But if we know where these trajectories are likely to end up, by modelling the transitions, then we can skip directly over the Worker's behaviour and instead follow the policy gradient of the predicted transition. 
FuN assumes a particular form for the transition model: that the direction in state-space, $s_{t+c} - s_t$, follows a von Mises-Fisher distribution. Specifically, if the mean direction of the von Mises-Fisher distribution
is given by $g(o_t)$ (which for compactness we write as $g_t$) we would have
$p(s_{t+c}|s_t,o_t) \propto e^{d_{\cos}(s_{t+c} - s_t, g_t)}$. If this functional form
\emph{were} indeed correct, then we see that our proposed update heuristic for the
Manager, eqn.\ref{eqn:manager_update}, is in fact the proper form for the transition
policy gradient arrived at in eqn.\ref{eqn:tpg}.

Note that the Worker's intrinsic reward (eqn.~\ref{eqn:intrinsic_reward}) is based on the log-likelihood of state trajectory. Through that the FuN architecture actively encourages the functional form of the transition model to hold true. Because the Worker is learning to achieve the Manager's direction, its transitions should, over time, closely follow a distribution around this direction, and hence our approximation for transition policy gradients should hold reasonably well. 

\seckiny
\section{Architecture details}
\label{sec:architecture}

This section provides the particular details of the model as described in section~\ref{sec:model}. The perceptual module $f^\textit{percept}$ is a convolutional network (CNN) followed by a fully connected layer. The CNN has a first layer with 16 8x8 filters of stride 4, followed by a layer with with 32  4x4 filters of stride 2. The fully connected layer has 256 hidden units.  Each convolutional and fully-connected layer is followed by a rectifier non-linearity\footnote{This is substantially the same CNN as in~\cite{mnih2016asynchronous,mnih-dqn-2015}, the only difference is that in the pre-processing stage we retain all colour channels.}. The state space which the Manager implicitly models in formulating its goals is computed via $f^\textit{Mspace}$, which is another fully connected layer followed by a rectifier non-linearity. The dimensionality of the embedding vectors, $w$, is set as $k=16$. To encourage exploration in transition policy, at every step with a small probability $\epsilon$ we emit a random goal sampled from a uni-variate Gaussian.

The Worker's recurrent network $f^\textit{Wrnn}$ is a standard LSTM~\cite{hochreiter1997lstm}. For the Manager's recurrent network,  $f^\textit{Mrnn}$, we propose a novel  design -- the dilated LSTM, which is introduced in the next section. Both  $f^\textit{Mrnn}$ and $f^\textit{Wrnn}$ have $256$ hidden units.

\begin{figure*}[th]
\subfloat[]{ \adjustbox{trim={.1\width} {.6\height} {.7\width} {.07\height},clip}%
{\includegraphics[scale=0.5]{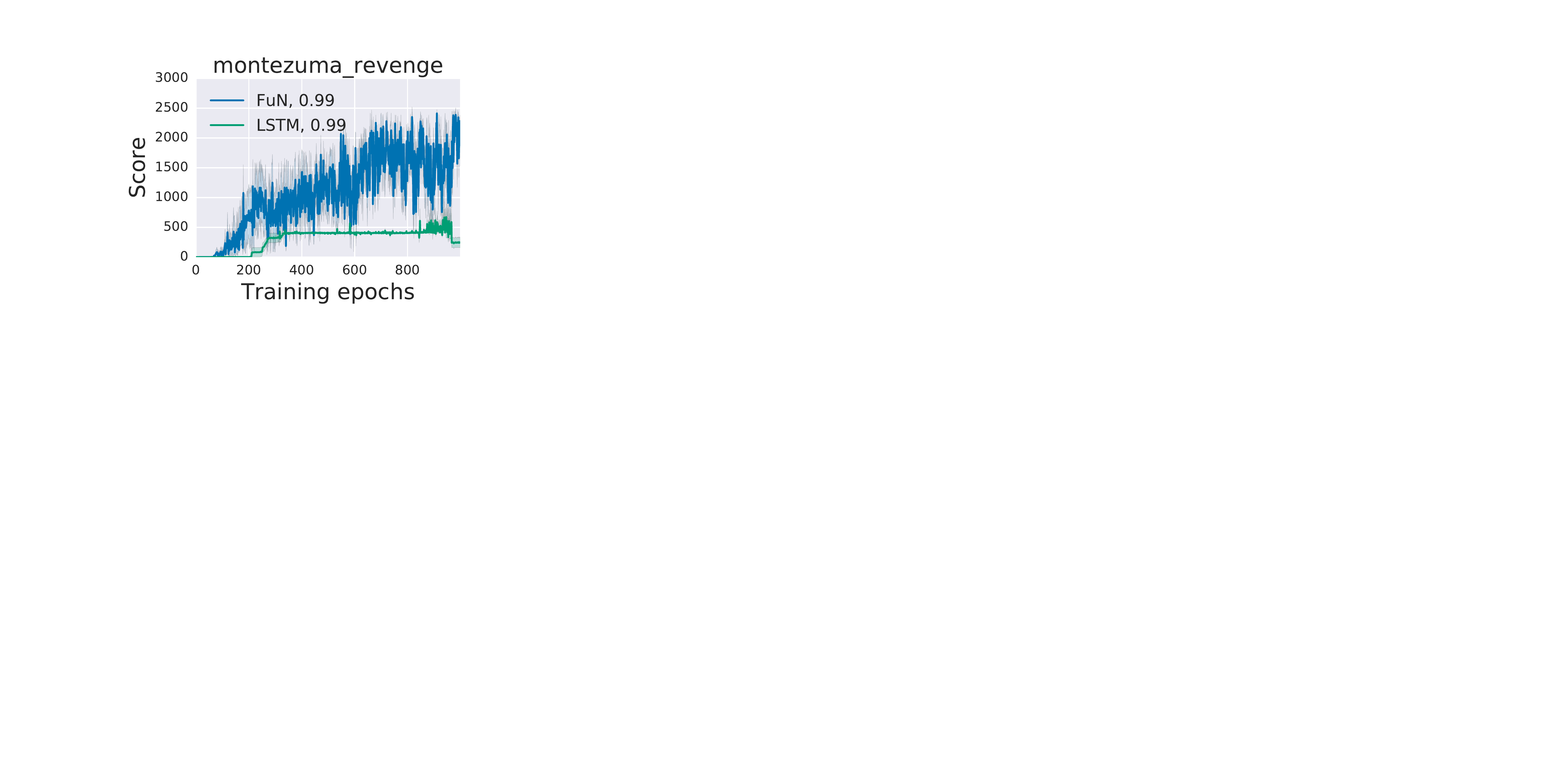}}  } \hspace{25mm}
\subfloat[]{ \includegraphics[scale=0.6]{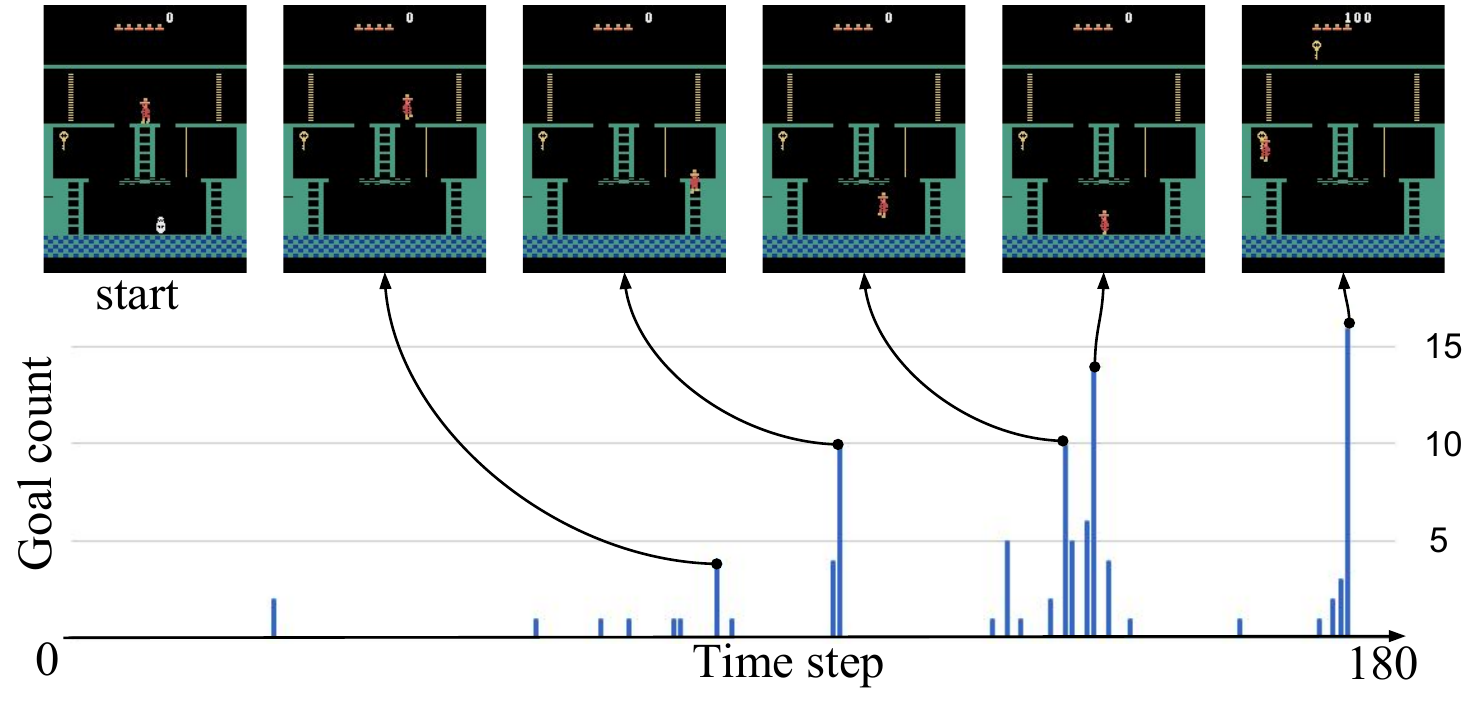} }
\caption{a) Learning curve on Montezuma's Revenge b) This is a visualisation of sub-goals learnt by FuN in the first room. For each time step we compute the latent state $s_t$ and the corresponding goal $g_t$. We then find a future state for which $\cos(s_{t’} - s_t, g_t)$ is maximized. The plot corresponds to the number of past states for which a frame maximizes the goal - i.e. the taller the bar, the more frequently that state was a maximizer of the expression for some previous state. Notice that FuN has learnt a semantically meaningful sub-goals -- the tall bars in the plot (i.e. consistent goals) correspond to interpretably useful “waypoints” in Montezuma.\label{fig:Montezuma} \figskiny }
\end{figure*}

\subseckiny
\subsection{Dilated LSTM}
\subseckiny
\label{subsec:dLSTM}
We propose a novel RNN architecture for the Manager, which operates at lower temporal resolution than the data stream. We define a dilated LSTM analogously to dilated convolutional networks~\cite{yu2015dilated}. For a dilation radius $r$ let the full state of the network be $h = \{\hat{h}^i\}_{i=1}^r$, i.e. it is composed of $r$ separate groups of sub-states or `cores'. At time $t$ the network is governed by the following equations:
$\hat{h}^{t\%r}_t,g_t = \textit{LSTM}(s_t, \hat{h}^{t\%r}_{t-1}; \theta^{\textit{LSTM}})$,
where $\%$ denotes the modulo operation and allows us to indicate which group of cores is currently being updated. 
We make the parameters of the LSTM network $\theta^{\textit{LSTM}}$ explicit to stress that the same set of parameters governs the update for each of the $r$ groups within the dLSTM. 

At each time step only the corresponding part of the state is updated and the output is pooled across the previous $c$ outputs. This allows the $r$ groups of cores inside the dLSTM to preserve the memories for long periods, yet the dLSTM as a whole is still able to process and learn from every input experience, and is also able to update its output at every step. 
This idea is similar to clockwork RNNs~\cite{koutnik2014clockwork}, however there the top level ``ticks'' at a fixed, slow pace, whereas the dLSTM observes all the available training data instead. In the experiments we set $r=10$, and this was also used as the predictions horizon, $c$.
\seckiny

\begin{figure*}[h!]
\includegraphics[scale=0.7]{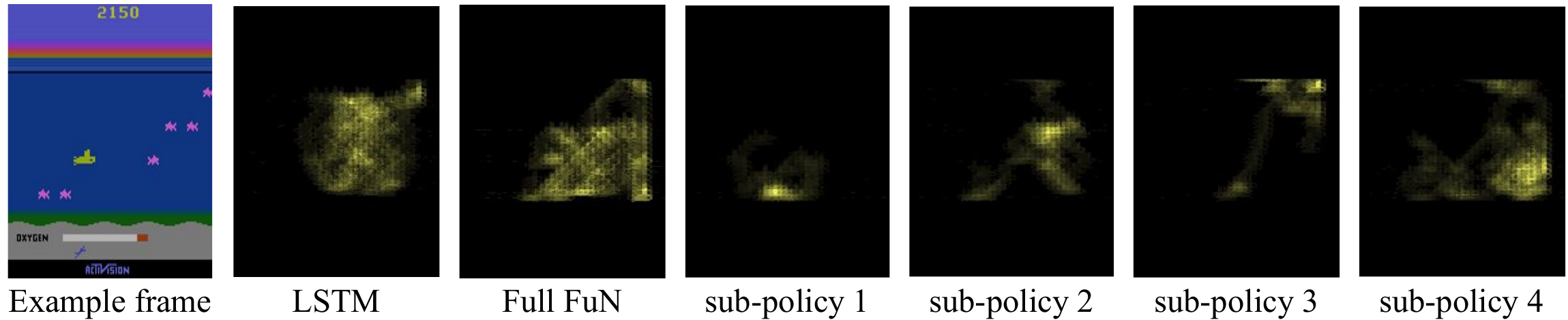}
  \caption{Visualisation of sub-policies learnt on sea quest game. We sample a random goal and feed it as a constant conditioning for the Worker and record its behaviour. We filter out only the image of the ship and average the frames, acquiring the heat-map of agents spatial location. From left to right: i) an example frame of the game ii) policy learnt by LSTM baseline iii) full policy learnt by FuN followed by set of different sub-policies. Notice how sub-policies are concentrated around different areas of the playable space. Sub-policy 3 is used to swim up for oxygen. \label{fig:SeaQ_subgoals} \figskiny }
\end{figure*}

\section{Experiments}
\seckiny
\label{sec:Experiments}

The goal of our experiments is to demonstrate that FuN learns non-trivial, helpful, and interpretable sub-policies and sub-goals, and also to validate components of the architecture. 
We start by describing technical details of the experimental setup and then present results on Montezuma's revenge -- an infamously hard ATARI game -- in section~\ref{subsec:Montezuma}. 
Section~\ref{subsec:atari} presents results on more ATARI games and extensively compares FuN to LSTM baseline with different discount factors and BPTT lengths.
In section~\ref{subsec:Mem_in_Lab} we present results on a set of visual memorisation tasks in 3D environment. 
Section~\ref{subsec:ablative} presents an ablation study of FuN, validating our design choices.

\parskiny
\paragraph{Baseline.} Our main baseline is a recurrent LSTM network on top of a representation learned by a CNN. The LSTM~\cite{hochreiter1997lstm} architecture is a widely used recurrent network and it was demonstrated to perform very well on a suite of reinforcement learning problems~\cite{mnih2016asynchronous}. LSTM uses 316 hidden units\footnote{This choice means that FuN and the LSTM baseline to have roughly the same number of total parameters.} and its inputs are the feature representation of an observation and the previous action of the agent.  Action probabilities and the value function estimate are regressed from its hidden state. All the methods the same CNN architecture, input pre-processing, and an action repeat of 4.

\begin{figure*}[th]
{\includegraphics[scale=0.28]{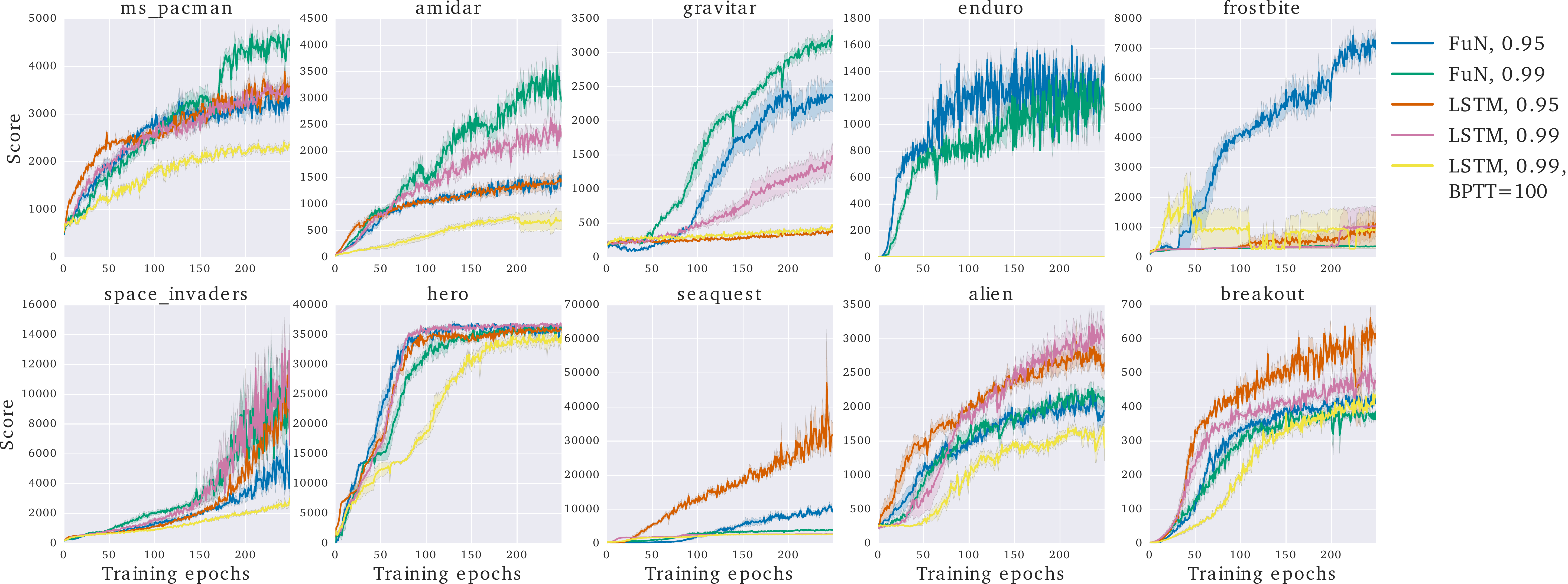}}
  \caption{ ATARI training curves. Epochs corresponds to a million training steps of an agent. The value is the average per episode score of top 5 agents, according to the final score. We used two different discount factors $0.95$ and $0.99$. \label{fig:ATARI_all_curves} \figskiny }
\end{figure*}

\parskiny
\paragraph{Optimisation.} We use the A3C method~\cite{mnih2016asynchronous} for all reinforcement learning experiments. It was shown to achieve state-of-the-art results on several challenging benchmarks~\cite{mnih2016asynchronous}.
We cut the trajectory and run backpropagation through time (BPTT)~\cite{mozer1989focused} after $K$ forward passes of a network or if a terminal signal is received. For FuN $K=400$, for LSTM, unless otherwise stated, $K=40$. We discuss different choice of $K$ for LSTM in section~\ref{subsec:atari}.
The optimization process runs 32 asynchronous threads using shared RMSProp. There are 3 hyper-parameters in FuN and 2 in the LSTM baselines. For each method, we ran 100 experiments, each using randomly sampled hyper-parameters. Learning rate and entropy penalty were sampled from a $\mathit{LogUniform}(10^{-4},10^{-3})$ interval for LSTM. For FuN the learning rate was sampled from $\mathit{LogUniform}(10^{-4.5},10^{-3.5})$, to account for higher gradients due to longer BPTT unrolls. The learning rate was linearly annealed from a sampled value to half the initial rate for all agents. To explore intrinsic motivation in FuN, we sample its weight $\alpha \sim \mathit{Uniform}(0,1)$. We define a training epoch as one million observations.  When reporting learning curves, we plot the average episode score of the top 5 agents (according to the final score) against the training epochs. For all ATARI experiments we clip the reward to $[-1,+1]$ interval

\subseckiny
\subsection{Montezuma's revenge}
\subseckiny
\label{subsec:Montezuma}

Montezuma's revenge is one of the hardest games available through the ALE~\cite{bellemare-ale}. The game is infamous for challenging agents with lethal traps and sparse rewards.  We had to broaden and intensify our hyper-parameter search for the LSTM baseline to see any progress at all for that model. We have experimented with many different hyper-parameter configurations for LSTM baseline, for instance expanding learning rate search to $\mathit{LogUniform}(10^{-3},10^{-2})$, and we report on the configuration that worked best. We use a small discount $0.99$ for LSTM; for FuN we use $0.99$ in Worker and $0.999$ in Manager.
Figure~\ref{fig:Montezuma}b analyses the sub-goals learnt by FuN in the first room. They turn out to be meaningful milestones, which bridge the agents progress to its first extrinsic reward -- picking up the key. Interestingly, two of the learnt sub-goals correspond to roughly the same locations as the ones hand-crafted in~\cite{tejas2016hdrl} (ladder and key), but here they are learnt by the agent itself. Figure~\ref{fig:Montezuma}a plots the learning curves. Notice how FuN starts learning much earlier and achieves much higher scores. It takes $>300$ epochs for LSTM to reach the score $400$, which corresponds to solving the first room (take the key, open a door); it stagnates at that score until about $900$ epochs, when it starts exploring further. FuN solves the first room in less than $200$ epochs and immediately moves on to explore further, eventually visiting several other rooms and scoring up to $2600$ points.

\begin{figure*}
\subfloat[]
{\includegraphics[scale=0.5]{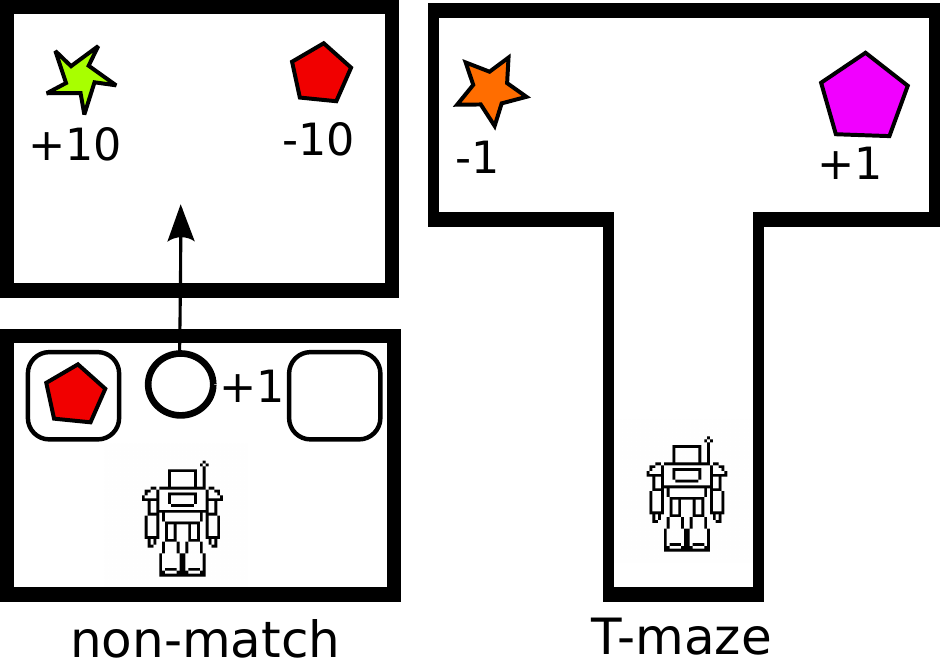}  } 
\subfloat[]{ \includegraphics[scale=0.5]{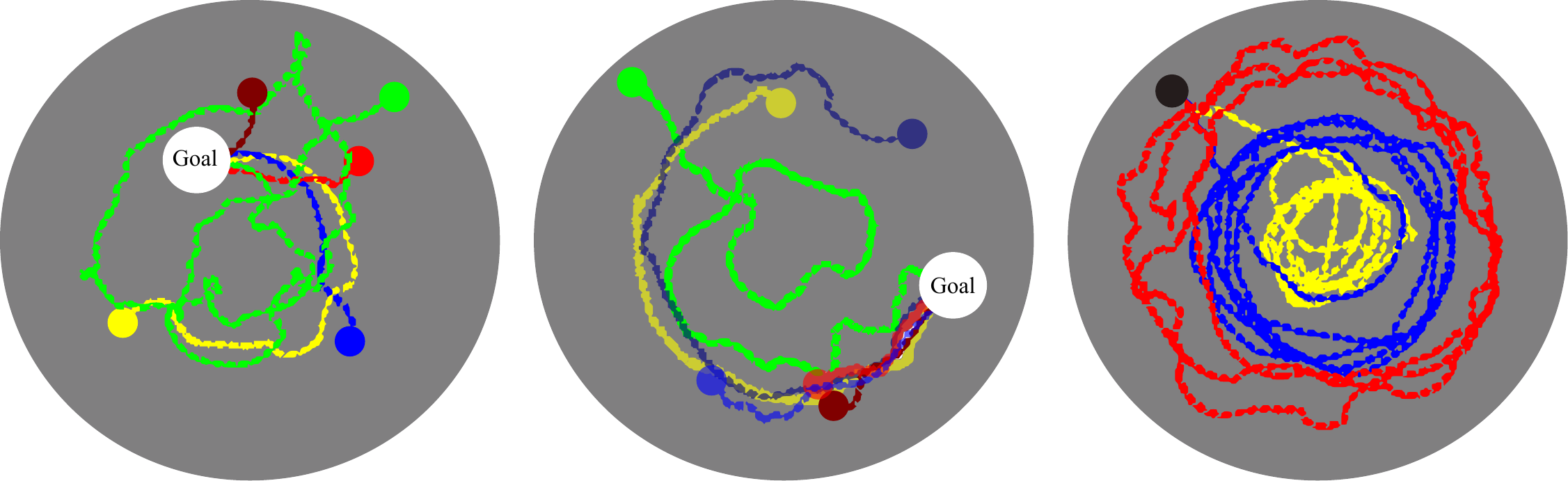} }
\caption{ a) Schematic illustration of t-maze and non-match domains b) FuN in water maze. First two plots from the left, are a visualisation of FuN trajectories during one episode. The first trajectory (green) performs a search for the target in different locations, while subsequent ones (other colours) perform searches along a circle of a fixed radius matched to that of the target, always finding the target. The rightmost plot visualises different learnt sub-policies, produced by sampling a random $g$ and fixing it for 200 steps. Each colour corresponds to a different $g$, the black circle represents the starting location. \label{fig:watermaze_options} \figskiny  }
\end{figure*}

\begin{figure*}[h]
\adjustbox{trim={.05\width} {.48\height} {0.05\width} {.068\height},clip}%
{\includegraphics[scale=0.35]{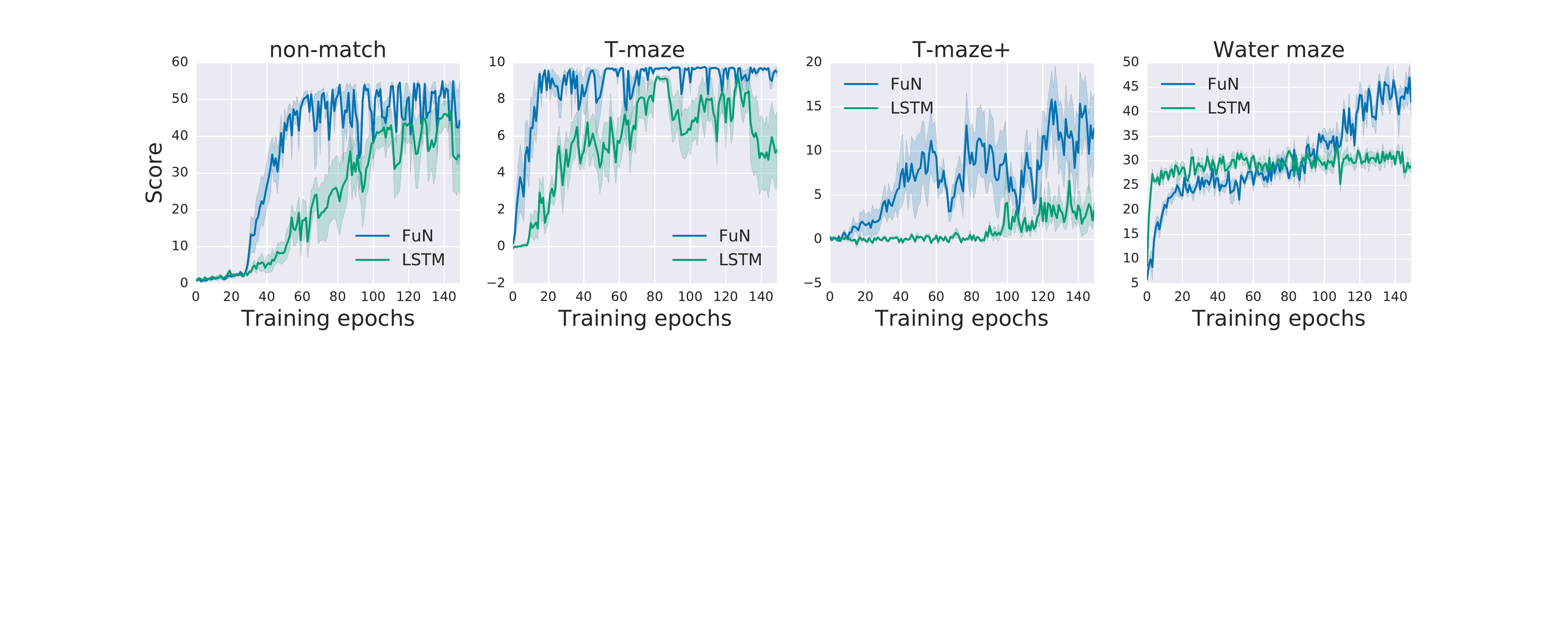}}
  \caption{ Training curves for memory tasks on Labyrinth. \label{fig:Laby_all_curves} \figskiny }
\end{figure*}

\subseckiny
\subsection{ATARI}
\subseckiny
\label{subsec:atari}

Experiments in this section validate that the capabilities of FuN go beyond what standard tools for long-term credit assignment -- discount factors and BPTT unroll length -- can provide for a baseline LSTM agent.
We use two discounts $0.99$ and $0.95$ for both FuN and LSTM agents. (For the experiments on FuN only the discount for the Manager changes, while the Worker's discount is fixed at $0.95$.) For the LSTM we explore BPTT of $40$ and $100$, while for FuN we use a BPTT unroll of $400$. For LSTM with BPTT $100$ we search for learning rate in the interval  $\mathit{LogUniform}(10^{-4.5},10^{-3.5})$, as for FuN. We use a diverse set of ATARI games, some of which involve long-term credit assignment and some which are more reactive. 

Figure~\ref{fig:ATARI_all_curves} plots the learning curves. A few categories emerge. On Ms. Pacman, Amidar, and Gravitar FuN with a low Manager discount of $0.99$ strongly outperforms all other methods. All of these games are known to require long-term reasoning to play well. Enduro stands out as all the LSTM agents completely fail at it. In this game the agent controls a racing car and scores points for overtaking other racers; this requires accelerating and steering for significant amount of time before the first reward is experienced. Frostbite is a hard game~\cite{vezhnevets2016straw, lake2016likehumans} that requires both long-term credit assignment and good exploration. The best-performing frostbite agent is FuN with $0.95$ Manager discount, which outperforms the rest by a factor of 7. On Hero and Space Invaders all agents perform equally well. 
On Seaquest and Breakout, the baseline LSTM with a more aggressive discount of $0.95$ is the best. This suggests that in these games long-term credit assignment is not important and the agent is better off optimising more immediate rewards in a greedy fashion. Alien is the only game where using different discounts doesn't meaningfully influence the agents performance; here we see the baseline LSTM outperforms our FuN model, although both still achieve a satisfactory scores. We provide qualitative analysis of sub-policies learnt on Seaquest in supplementary material.

Note how using an unroll for BPTT=100 in the baseline LSTM significantly hurts its performance (hence we do not explore longer unrolls), while FuN performs very well with BPTT of $400$ thanks to its ability to leverage the dLSTM. 
Being able to train a recurrent network over very long sequences could be an enabling tool for many memory related task, as we demonstrate in section~\ref{subsec:Mem_in_Lab}. 

\paragraph{Qualitative analysis on Seaquest}

To qualitatively inspect sub-policies learnt by the Worker we use the following procedure: first, we record goals emitted by Manager during the play; we then sample one of them and provide it as a constant input to the Worker for the duration of an episode and record its behaviour. This allows us to qualitatively inspect what kind of sub-policies emerge. Figure~\ref{fig:SeaQ_subgoals} plots sub-policies learnt on the seaquest game. Notice how different options correspond to rough spatial positions or manoeuvres for the agent's submarine -- for instance sub-policy 3 corresponds to swimming up for air.

\begin{figure}[h]
\includegraphics[scale=0.35]{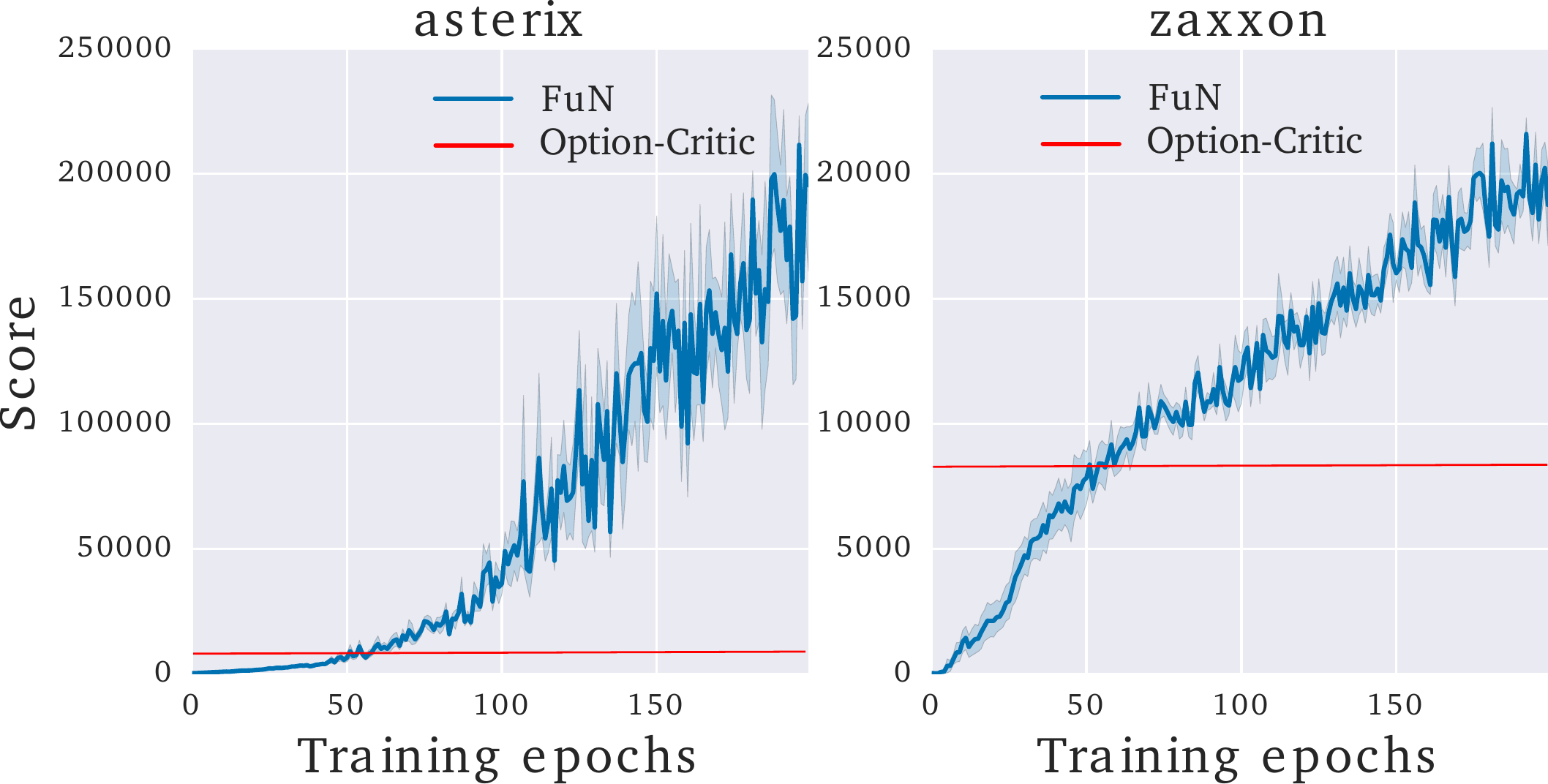}
  \caption{Comparison to Option-Critic on Zaxxon and Asterix. Score for Option-Critic is taken from the original paper  \label{fig:OptCrit} \figskiny }
\end{figure}

\parskiny
\paragraph{Option-critic architecture}~\cite{bacon2015option} is, to the best of our knowledge, the only other end-to-end trainable system with sub-policies. The experimental results for Option-Critic on 4  ATARI~\cite{bacon2015option} games show scores similar those from a flat DQN~\cite{mnih-dqn-2015} baseline agent. Notice that our baseline~\cite{mnih2016asynchronous} is much stronger than DQN. We also ran FuN on the same games as Option-Critic (Asterix, Ms. Pacman, Seaquest and Zaxxon) and after 200 epochs it achieves a similar score on Seaquest, doubles it on Ms. Pacman, more than triples it on Zaxxon and gets more than 20x improvement on Asterix.  Figure~\ref{fig:OptCrit} presents our results on Asterix and Zaxxon. We took the approximate performance of Option-Critic from the original paper -- 8000 for Asterix and 6000 for Zaxxon. Plots in the original paper also suggest that score stagnates around these levels, notice that our score keeps going up.

\begin{figure*}[h]
\includegraphics[scale=0.35]{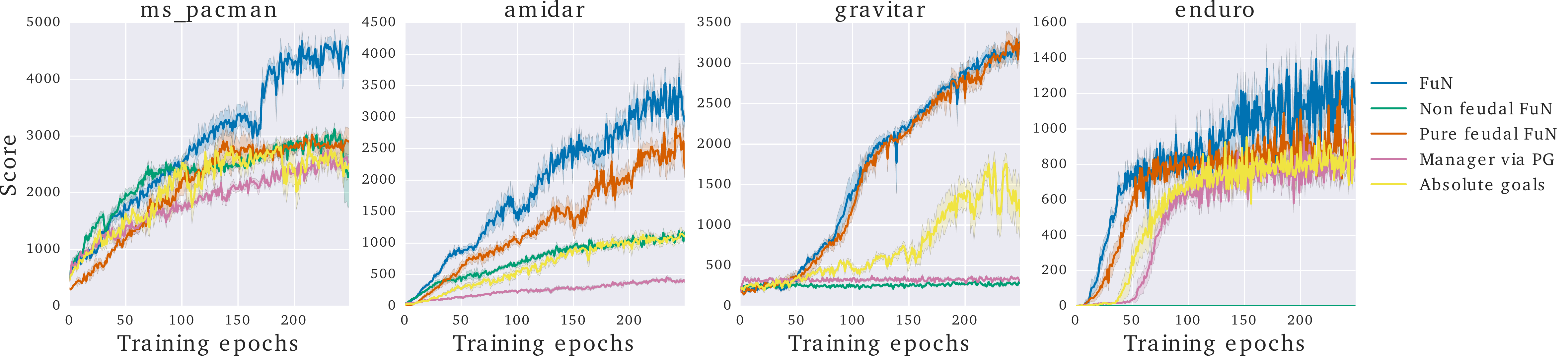}
  \caption{ Ablative analysis \label{fig:Ablation_curves} }
\end{figure*}

\begin{figure*}[h]
{\includegraphics[scale=0.35]{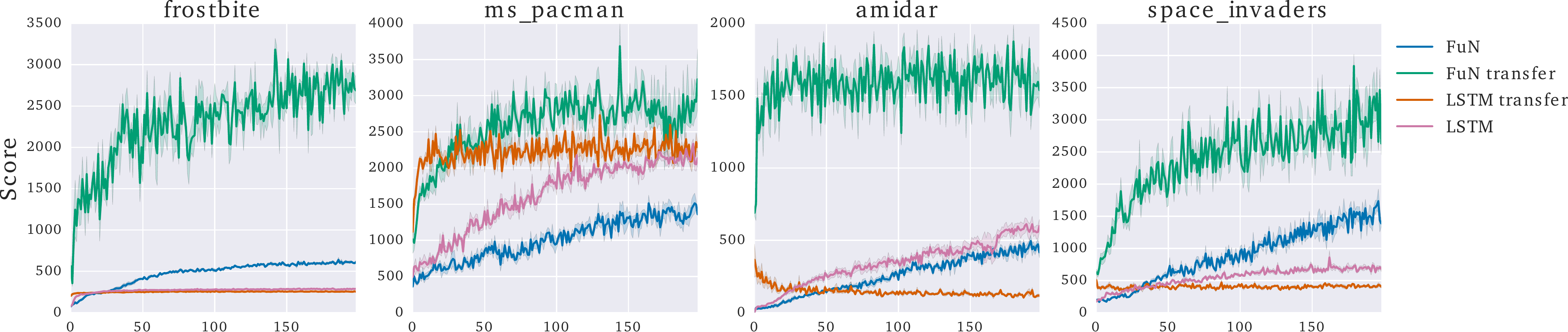}}
  \caption{ Action repeat transfer \label{fig:ActRep_graphs} \figskiny }
\end{figure*}

\subseckiny
\subsection{Memory in Labyrinth}
\subseckiny
\label{subsec:Mem_in_Lab}
DeepMind Lab~\cite{beattie2016labyrinth} is a first-person 3D game platform extended from OpenArena. It's a visually complex 3D environment with agent actions corresponding to movement and orientation. We use 4 different levels that test long-term credit assignment and visual memory: 

\parskiny
\paragraph{Water maze} is a reproduction of the Morris water maze experiment~\cite{morris1981watermaze} from the behavioural science literature. An agent is dropped into a circular pool of water with a concealed platform at unknown random location. The agent can move around and upon stepping on the platform it receives a reward and the trial restarts. The platform remains in the same location for the rest of the episode, while agent starts each trial at a random location. The walls of the pool are decorated with visual cues to assist localisation. 

\parskiny
\paragraph{T-maze} is another classic animal cognition test. The agent spawns in a small T-shaped maze.  Two objects with randomly chosen shape and colour are spawned at the left and right "baiting" locations.  One of them is assigned a reward of +1 and the other a reward of -1.  When the agent collects one of the objects, it receives the reward and is re-spawned at the beginning of the T-maze. The objects are also re-instantiated in the same locations and with the same rewards on the re-spawn event. The agent should remember which object gives the positive reward across re-spawns and collect it as many times as possible within the fixed time given for the episode. \textbf{T-maze+} is a modification of T-maze, where at each trial the length of corridors can vary, adding additional dimension of complexity. 

\parskiny
\paragraph{Non-match} is a visual memorisation task. Each trial begins in small room with an out of reach object being displayed in one of two display pods. There is a pad in the middle, which upon touching, the agent is rewarded with 1 point, and is teleported to a second room which has two objects in it, one of which matches the object in the previous room. Collecting the matching object gives a reward of -10 points, collecting the non matching object gives a reward of 10 points. Once either is collected, the agent is teleported back to the first room, with the same object being shown.

For all agents we include reward as a part of the observation.
Figure~\ref{fig:watermaze_options}a illustrates T-maze and non-match environments and 
figure~\ref{fig:Laby_all_curves} plots the learning curves. FuN consitently outperforms the LSTM baseline -- it learns faster and also reaches a higher final reward. We analyse the FuN agent's behaviour in more detail in Figure~\ref{fig:watermaze_options}b. It demonstrates that FuN learns meaningful sub-policies, which are then efficiently integrated with memory to produce rewarding behaviour. Interestingly, the LSTM agent doesn't appear to use its memory for water maze task at all, always circling the maze at the roughly the same radius. 

\subseckiny
\subsection{Ablative analysis}
\subseckiny
\label{subsec:ablative}

This section empirically validates the main innovations of this paper: transition policy gradient for training the Manager; relative rather than absolute goals; lower temporal resolution for Manager; intrinsic motivation for the Worker.

\paragraph{Transition policy gradient} First we consider a `non-Feudal' FuN -- it has exactly the same network architecture as FuN, but the Manager’s output $g$ is trained with gradients coming directly from the Worker and no intrinsic reward is used, much like in Option-Critic architecture~\cite{bacon2015option}.  Second, $g$ is learnt using a standard policy gradient approach with the Manager emitting the mean of a Gaussian distribution from which goals are sampled (as if the Manager were solving a continuous control problem~\cite{schulman2015gae,mnih2016asynchronous,lillicrap2015continuous}). Third, we explore a variant of FuN in which $g$ specifies absolute, rather than relative/directional, goals (and the Worker's intrinsic reward is adjusted accordingly) but otherwise everything is the same.
The experiments (Figure~\ref{fig:Ablation_curves}) reveal that, although alternatives do work to some degree their performance is significantly inferior. We also evaluate a purely feudal version of FuN -- in which the Worker is trained from the intrinsic reward alone. This ablation performs better than other, but still inferior to the full FuN approach. It shows that allowing the Worker to experience the external reward is beneficial. 

\begin{figure*}[h!]
\includegraphics[scale=0.35]{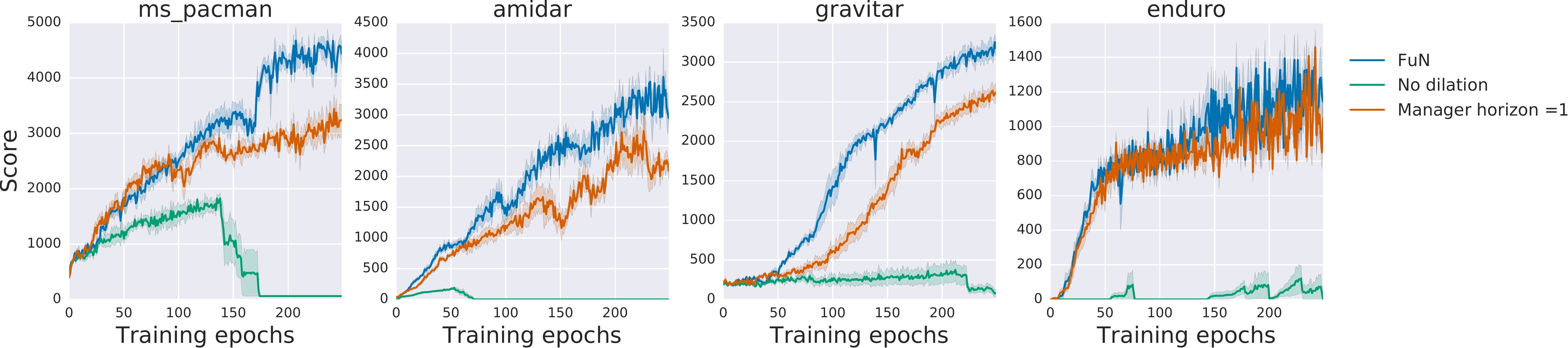}
  \caption{ Learning curves for ablations of FuN that investigate influence of dLSTM in the Manager and Managers prediction horizon $c$. No dilation -- FuN trained with a regular LSTM in the Manager; Manager horizon =1 -- FuN trained with $c=1$.\label{fig:ablation_temp} \figskiny }
\end{figure*}

\paragraph{Temporal resolution ablations.}  An important feature of FuN is the ability of the Manager to operate at a low temporal resolution. This is achieved through dilation in the LSTM and through the prediction horizon $c$. To investigate their influence we use two baselines:
i) the Manager uses a vanilla LSTM with no dilation; ii) FuN with Manager's prediction horizon $c=1$. Figure~\ref{fig:ablation_temp} presents the results. The non-dilated LSTM fails catastrophically, most likely overwhelmed by the recurrent gradient. Reducing the horizon $c$ to 1 did hurt the performance, although interestingly less so than other ablations. It seems that even at high temporal resolution Manager captures certain properties of the underlying MDP and communicate them down to Worker in a helpful way. This confirms that by learning in two separate formulations FuN is able to capture richer structural properties of the environment and thus train faster. 

\begin{figure*}[h!]
\includegraphics[scale=0.22]{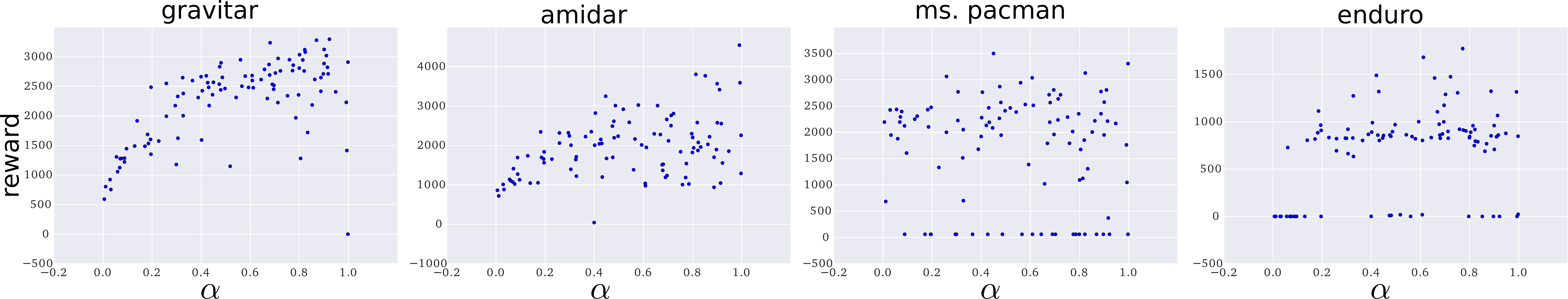}
  \caption{Scatter plot of agents reward after $200$ epochs vs intrinsic reward weight $\alpha$. \label{fig:alpha_scatter_plots} \figskiny }
\end{figure*}

\paragraph{Intrinsic motivation weight.} This section look at the impact of the weight $\alpha$, which regulates the relative weight of intrinsic reward (if $\alpha=0$ then intrinsic reward is not used). We train agents with learning rate and entropy penalty fixed to $10^{-3.5}$ and only vary $\alpha$ between $[0,1]$. Figure~\ref{fig:alpha_scatter_plots} shows scatter plots of agent’s final score vs $\alpha$ hyper-parameter.  Notice a clear correlation between the score and high value of $\alpha$ on gravitar and amidar; however on other games the optimal value of $\alpha$ can be less than to $1$.

\begin{figure*}[h!]
\includegraphics[scale=0.25]{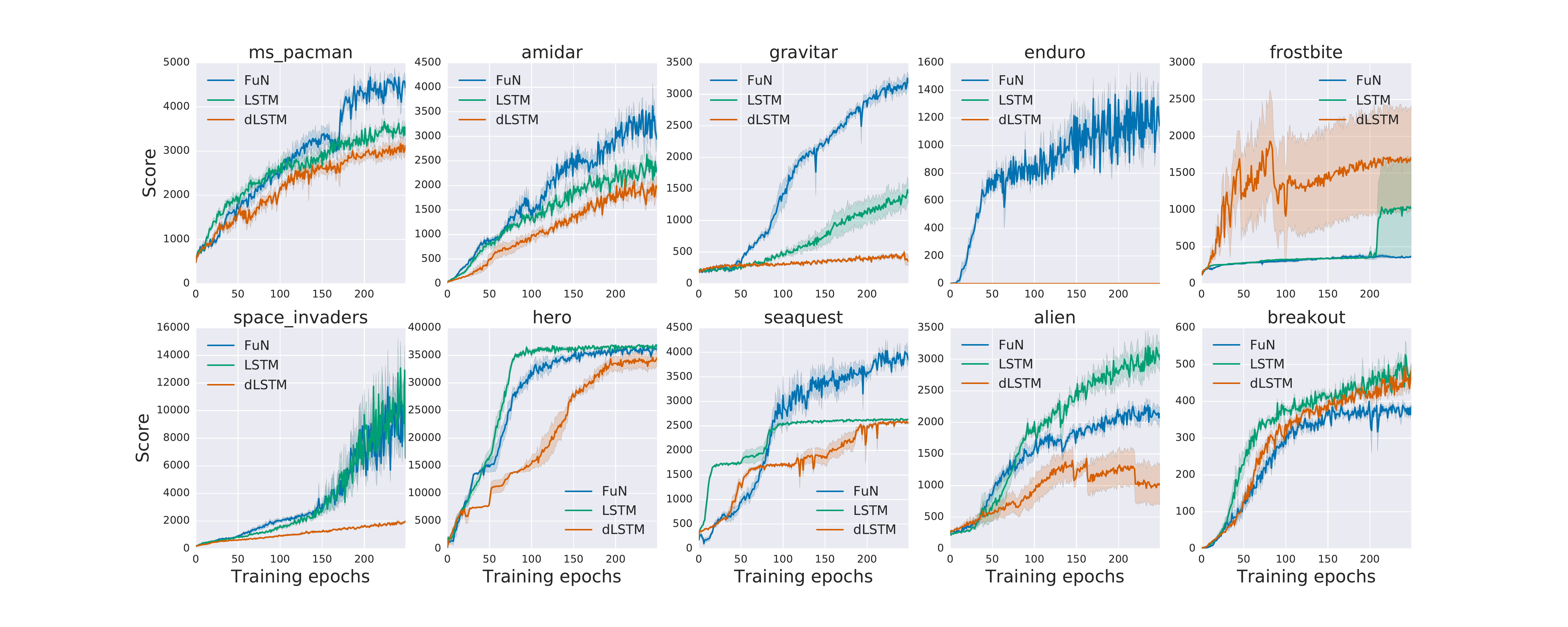}
  \caption{Learning curves for dLSTM based agent with LSTM and FuN for comparison. \label{fig:dLSTM_curves} \figskiny }
\end{figure*}

\paragraph{Dilate LSTM agent baseline}
One of innovations this paper presents is dLSTM design for a Recurrent network. In principle, it could alone be used in an agent on top of a CNN, without the rest of FuN structures. We evaluate such an agent as an additional baseline. We use the same hyper-parameters as for FuN -- BPTT=400, discount $=0.99$, learning rate sampled in the interval $\mathit{LogUniform}(10^{-4.5},10^{-3.5})$, entropy penalty $\mathit{LogUniform}(10^{-4},10^{-3})$. Figure~\ref{fig:dLSTM_curves} plots the learning curves for FuN, LSTM and dLSTM agents. dLSTM generally underperforms both LSTM and FuN. The power of dLSTM is in the ability to operate at lower temporal resolution, which is useful in the Manager, but not so much on it's own.

\subseckiny
\subsection{ATARI action repeat transfer}
\subseckiny
One of the advantages of FuN is the clear separation of duties between Manager and Worker. The Manager learns a transition policy, while the Worker learns to operate primitive actions to enact these transitions. This transition policy is invariant to the underlying embodiment of the agent -- the way its primitive actions translate into state space transitions. Potentially, the transition policy can be transferred between agents with different embodiment -- e.g. robot models with physical design or different operational frequency. We provide evidence towards that possibility by transferring policies across agents with different action repeat on ATARI.\footnote{Action repeat is a heuristic used in all successful agents~\cite{mnih-dqn-2015, mnih2016asynchronous, bellemare2016gap, vezhnevets2016straw}. It enables better exploration, eases credit assignment, and saves computation by repeating any action chosen by the agent several ($=4$) times.}

To perform transfer, we initialise the FuN system with parameters extracted from an agent trained with action repeat of $4$ and then make the following adjustments: (i) we accordingly adjust the discounts for all rewards; (ii) we increase the dilation of the dLSTM by a factor of 4; (iii) we increase the Manager's goal horizon $c$ by a factor of 4. (These modifications adapt all the ``hard-wired'' but explicitly temporally sensitive aspects of the agent.) We then train this agent without action repeat. As a baseline we use an LSTM agent transferred in a similar way (with adjusted discounts) as well as FuN and LSTM agents trained without action repeat from scratch. Figure~\ref{fig:ActRep_graphs} shows the corresponding learning curves. The transferred FuN agent (green curve) significantly outperforms every other method. Furthermore it shows positive transfer on each environment, whereas LSTM only shows positive transfer on Ms. Pacman.

\seckiny
\seckiny
\section{Discussion and future work}
\label{sec:conclusion}
\seckiny

How to create agents that can learn to decompose their behaviour into meaningful primitives and then reuse them to more efficiently acquire new behaviours is a long standing research question. 
The solution to this question may be an important stepping stone towards agents with general intelligence and competence. %
This paper introduced FeUdal Networks, a novel architecture that formulates sub-goals as directions in latent state space, which, if followed, translate into a meaningful behavioural primitives. 
FuN clearly separates the module that discovers and sets sub-goals from the module that generates the behaviour through primitive actions. This creates a natural hierarchy that is stable and allows both modules to learn in complementary ways. 
Our experiments clearly demonstrate that this makes long-term credit assignment and memorisation more tractable. 
This also opens many avenues for further research, for instance: deeper hierarchies can be constructed by setting goals at multiple time scales, scaling agents to truly large environments with sparse rewards and partial observability.
The modular structure of FuN is also lends itself to transfer and multitask learning -- learnt behavioural primitives can be re-used to acquire new complex skills, or alternatively the transitional policies of the Manager can be transferred to agents with different embodiment.  

\section{Acknowledgements}
We thank Alex Graves, Daan Wierstra, Olivier Pietquin, Oriol Vinyals, Joseph Modayil and Vlad Mnih for many helpful
discussions, suggestions and comments on the paper.

\bibliography{deeprl}
\bibliographystyle{icml2016}

\end{document}